\documentclass[fleqn,12pt]{wlscirep}
\usepackage[utf8]{inputenc}
\usepackage{algpseudocode}
\usepackage[T1]{fontenc}
\usepackage{subcaption}
\usepackage{pdflscape}
\usepackage{adjustbox}
\usepackage{algorithm}
\usepackage{todonotes}
\usepackage{tabularx}
\usepackage{amsfonts}
\usepackage{booktabs}
\usepackage{dashrule}
\usepackage{graphicx}
\usepackage{multicol}
\usepackage{multirow}
\usepackage{tabularx}
\usepackage{xcolor}
\usepackage{lineno}
\usepackage{tikz}
\usepackage[
    justification=justified
]{caption}

\title{Enhancing Global Maritime Traffic Network Forecasting with Gravity-Inspired\\Deep Learning Models}

\author[1,3,$\bigstar$]{Ruixin Song}
\author[2,3,$\bigstar$]{Gabriel Spadon}
\author[3]{Ronald Pelot}
\author[2,4]{Stan Matwin}
\author[5,$\dagger$]{Amilcar Soares}
\affil[1]{~Department of Computer Science, Memorial University of Newfoundland, St. John's -- NL, Canada}
\affil[2]{~Institute for Big Data Analytics, Dalhousie University, Halifax -- NS, Canada}
\affil[3]{~Industrial Engineering Department, Dalhousie University, Halifax -- NS, Canada}
\affil[4]{~Institute of Computer Science, Polish Academy of Sciences, Warsaw, Poland}
\affil[5]{~Department of Computer Science and Media Technology, Linnaeus University, Växjö, Sweden}
\affil[$\bigstar$]{~These authors contributed equally to this work.}
\affil[$\dagger$]{~Corresponding author (amilcar.soares@lnu.se).}

\begin{abstract}
Aquatic non-indigenous species (NIS) pose significant threats to biodiversity, disrupting ecosystems and inflicting substantial economic damages across agriculture, forestry, and fisheries.
Due to the fast growth of global trade and transportation networks, NIS has been introduced and spread unintentionally in new environments.
This study develops a new physics-informed model to forecast maritime shipping traffic between port regions worldwide. The predicted information provided by these models, in turn, is used as input for risk assessment of NIS spread through transportation networks to evaluate the capability of our solution.
Inspired by the gravity model for international trades, our model considers various factors that influence the likelihood and impact of vessel activities, such as shipping flux density, distance between ports, trade flow, and centrality measures of transportation hubs.
Accordingly, this paper introduces transformers to gravity models to rebuild the short- and long-term dependencies that make the risk analysis feasible.
Thus, we introduce a physics-inspired framework that achieves an 89\% binary accuracy for existing and non-existing trajectories and an 84.8\% accuracy for the number of vessels flowing between key port areas, representing more than 10\% improvement over the traditional deep-gravity model.
Along these lines, this research contributes to a better understanding of NIS risk assessment.
It allows policymakers, conservationists, and stakeholders to prioritize management actions by identifying high-risk invasion pathways.
Besides, our model is versatile and can include new data sources, making it suitable for assessing international vessel traffic flow in a changing global landscape.
\end{abstract}

\begin{document}

\maketitle
\setcounter{secnumdepth}{1}
\section*{Introduction}
Globalization has rapidly increased marine shipping activities in the last decades.
According to a statistics report, container shipping has increased by 24 times in tonnage from 1980 to 2020~\cite{StatistaResearchGroup2022ContainerShipping}.
This significant growth has not only boosted global trade but has also raised environmental concerns, including increased ocean pollution and the introduction of \textcolor{black}{NIS} through ballast water across marine ecosystems~\cite{Barry2008BallastWater}.
Recent research has revealed a strong correlation between the introduction of NIS and the movement of ships through shipping networks.
These studies~\cite{Seebens2013RiskMarine, Kaluza2010ComplexNetwork} have utilized data from the Automatic Identification System (AIS), a location tracking system on ships that allows them to share their positions in real-time~\cite{haranwala2023data}.
This technology allows researchers to track individual/collective ships~\cite{spadon2023safer, spadon2022unfolding, DBLP:journals/tits/NguyenVHGF22, DBLP:journals/corr/abs-2109-03958}, predict small- or larger-scale shipping activities~\cite{ferreira2022semi, alam2024enhancing}, and assess the risk of introducing NIS through ballast water~\cite{Seebens2013RiskMarine} discharge.

AIS data has emerged as a promising source of information for studying maritime ship traffic patterns.
The mobility intrinsic in AIS data associated it with studies of mobility flows by using Origin-Destination (OD) models that combine physics with statistical mechanics, including the classical gravity model, which, inspired by Newton's {\it Law of Universal Gravitation}, measures the attractive force between two objects based on their masses and the distance between them~\cite{IsaacNewtonProposition75}.
The gravity-inspired OD models were introduced in early human mobility and migration studies~\cite{Zipf1946P1P2, Jung2008GravityModel}, where the number of commuters between two locations relies on population size and distances between origins and destinations as features.

The gravity theory describes the interaction between two entities proportionally to their masses and inversely proportional to their distance~\cite{IsaacNewtonProposition75}.
It permeates many areas of study that go beyond mobility and migration, such as the spreading of epidemics~\cite{VENTURA2022100544, Kramer2016SpatialSpreadEbola}, commercial trading~\cite{VanBergeijk2010GravityModel}, communication network~\cite{Gonzalez2008UnderstandingIndividual} and cargo shipping networks~\cite{Kaluza2010ComplexNetwork, Tu2018ShippingNetwork, Ducruet2020UrbanGravity} modeling.
Although the gravity model has been widely used to model real-world problems, recent studies have shown that it may not be sufficient for capturing complex patterns in various scenarios~\cite{Beyer2022GravityModels}.
Relying solely on mass and distance as the critical factors of the model could lead to failures in accurately representing patterns~\cite{Song2010LimitsPredictability}.
Also, the lack of limits on the flow size can lead to the predicted flow size larger than the source ``masses'', making the model predictions unreasonable and challenging to interpret~\cite{Simini2012UniversalModel, Spadon2019ReconstructingCommuters}.
Additionally, it is difficult to deal with multiple adjustable parameters in the model for the prediction without enough previously observed data.
Nonetheless, the gravity model has been prevalent for many years and remains a popular tool for modeling various phenomena.

Beyond gravity models, radiation absorption is another physics-inspired OD model to study mobility patterns~\cite{Simini2012UniversalModel, Masucci2013GravityRadiation, Ren2014PredictingCommuter}.
Unlike gravity models, radiation models are based on principles seen in radiation and absorption processes from physics.
The radiation model overcomes several limitations of gravity models in predicting mobility flows.
While gravity models contain adjustable variables that may be difficult to define, the radiation-absorption model simplifies this by emphasizing distance as the primary feature while considering the population density and offers a parameter-free approach that resolves the problem of having multiple parameters in the gravity model.
Besides, it limits the number of flows by introducing the total number of individuals departing from the source location~\cite{Simini2012UniversalModel, Ren2014PredictingCommuter}.
Further physics-inspired studies used field theory for abstracting vector fields of daily commuting flows~\cite{Mazzoli2019FieldTheory}, while others translated field theory-based mobility to deep learning models for achieving better interpretation of spatiotemporal features in mobility patterns~\cite{J.Wang2022TrafficFlow}.
Moreover, different studies have incorporated multiple relative factors, such as employment and urban infrastructure, that capture more important characteristics to improve prediction accuracy and adaptability of the model~\cite{Spadon2019ReconstructingCommuters, Simini2021DeepGravity}.
The evolving nature of artificial intelligence and the rise of large language models have brought about new technologies.
Therefore, improving the state of the art by using more capable technologies over new arrangements and combinations of data is essential.
This becomes more evident when considering evolving factors such as climate change and recurrent anthropogenic effects observed on the oceans.

Deep learning techniques have become increasingly popular in various applications due to their ability to recognize patterns by fine-tuning multiple parameters.
These techniques have been used to forecast vessel trajectories in the ocean, predict patient trajectories in hospitals, track the spread of epidemics, and many other applications~\cite{rodrigues2021ligdoctor, spadon2022evolution, s21124026}.
Understandably, coupling machine or deep learning capabilities for pattern recognition with physics-inspired OD models can imply a higher proficiency in capturing and predicting complex scenarios.
For instance, a study in human mobility used deep learning methods with the gravity model~\cite{Simini2021DeepGravity}.
The resulting composite model, called the Deep Gravity Model, has expanded the standard feature set of conventional gravity models, which typically incorporated population size and distance between OD locations, to include a variety of parameters characterizing the origins and destinations such as land-use patterns and the presence of retail and healthcare amenities.
However, the simplicity of the multilayer perceptrons (MLPs) structure in Deep Gravity models presents certain limitations.
Specifically, accurately capturing the complex multivariate relationships inherent in mobility flows can be challenging due to the composition of multiple functions, which provides an opportunity for further research efforts.

\begin{figure}[!b]
    \centering
    \includegraphics[width=.9\textwidth]{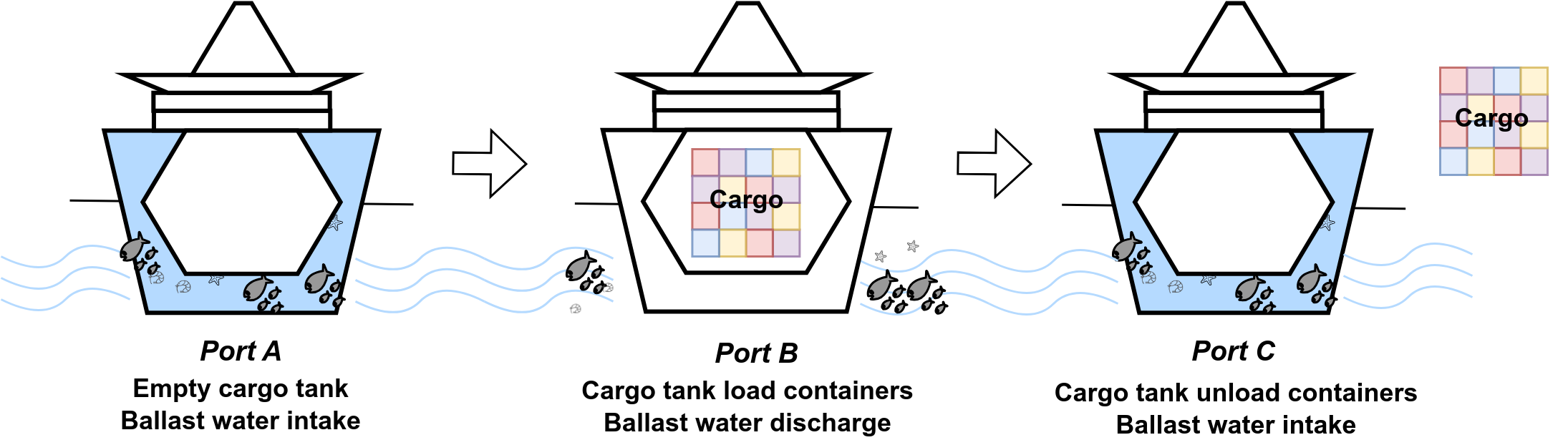} \\
    \caption{Non-indigenous species carried by ballast water during container shipping.}
    \label{fig:problem-example}
\end{figure}

Along these lines, this paper focuses on predicting the OD fluxes of marine vessels to gain insight into global shipping behavior and revert such information to the risk assessment of NIS cases through BWRA for performance comparison and evaluation against state-of-the-art machine-learning-driven solutions.
Ballast water is essential in maritime operations, particularly for container ships.
As illustrated in Figure~\ref{fig:problem-example}, while ballast water helps stabilize the ship's load, local species can enter the ballast tanks and travel long distances when the water is taken in and discharged.
Therefore, forecasting shipping patterns is critical to understanding the risk of spreading NIS.
To this end, we propose a gravity-informed model where the shipping fluxes are considered as ``mass'', and the vessel traffic flows are inversely proportional to the distances traveled by ships.
We have enriched our model with relevant features from shipping activities, including bilateral trade data between countries~\cite{Harvard2019InternationalTrade} and graph metrics extracted from the global shipping network.
Our deep learning model, known as the {\it Transformer Gravity}, relies on the transformer architecture~\cite{Vaswani2017AttentionAll} and is proficient in capturing local and global data dependencies through self-attention mechanisms.
This mechanism enables the model to weigh the importance of different parts of the sequence of input data ({\it i.e.}, information about maritime vessel movements), assigning varying degrees of attention depending on the relevance of each input part while generating the output.
As a result, the model we propose can better discern and incorporate short- and long-term dependencies in vessel traffic flows, making it more sensitive to the complex and dynamic patterns in maritime vessel movements.

As part of our proposed framework, we have employed a machine learning classifier that proceeds the flow estimation process by identifying the most probable OD pairs for gravity models. In this step, we first construct a fully connected shipping network to fill gaps in the original network. We weigh the edges based on distance and the low likelihood of a vessel traveling between the two locations (see Methods). The classifier then evaluates each link, learning how to filter out those less likely to exist based on critical attributes, including distance and edge importance (see Methods).
This allows only highly probable flows, based on prior knowledge, to be fed into the gravity-based models, where the final flow estimations take place with the aid of the gravity-based model.
In this sense, we have conducted experiments using the {\bf (a)} Transformer Gravity, {\bf (b)} Deep Gravity, and {\bf (c)} shallower-layered variants of Deep Gravity Models.
We utilized regression models in machine learning for performance comparison.
Our results demonstrate that the Transformer Gravity model significantly outperforms all the other approaches as it achieves an average Common Part of Commuters (CPC) of 86.4\%, representing an improvement higher than 10\% in the model output certainty in contrast to the benchmark Deep Gravity and its shallower-layered variants, and nearly 50\% improvement compared to machine learning regression models.
The results we have obtained are not only due to the proposed model but also to the pipeline outputs of the most probable connections that provide prior knowledge to the gravity-informed models (see Methods).
We have significantly improved performance by incorporating prior knowledge about potential destinations and using the attention mechanism and the traditional gravity-informed model for mobility flow estimation.

\section*{Results}
\begin{figure}[!t]
    \centering
    \includegraphics[width=.9\textwidth]{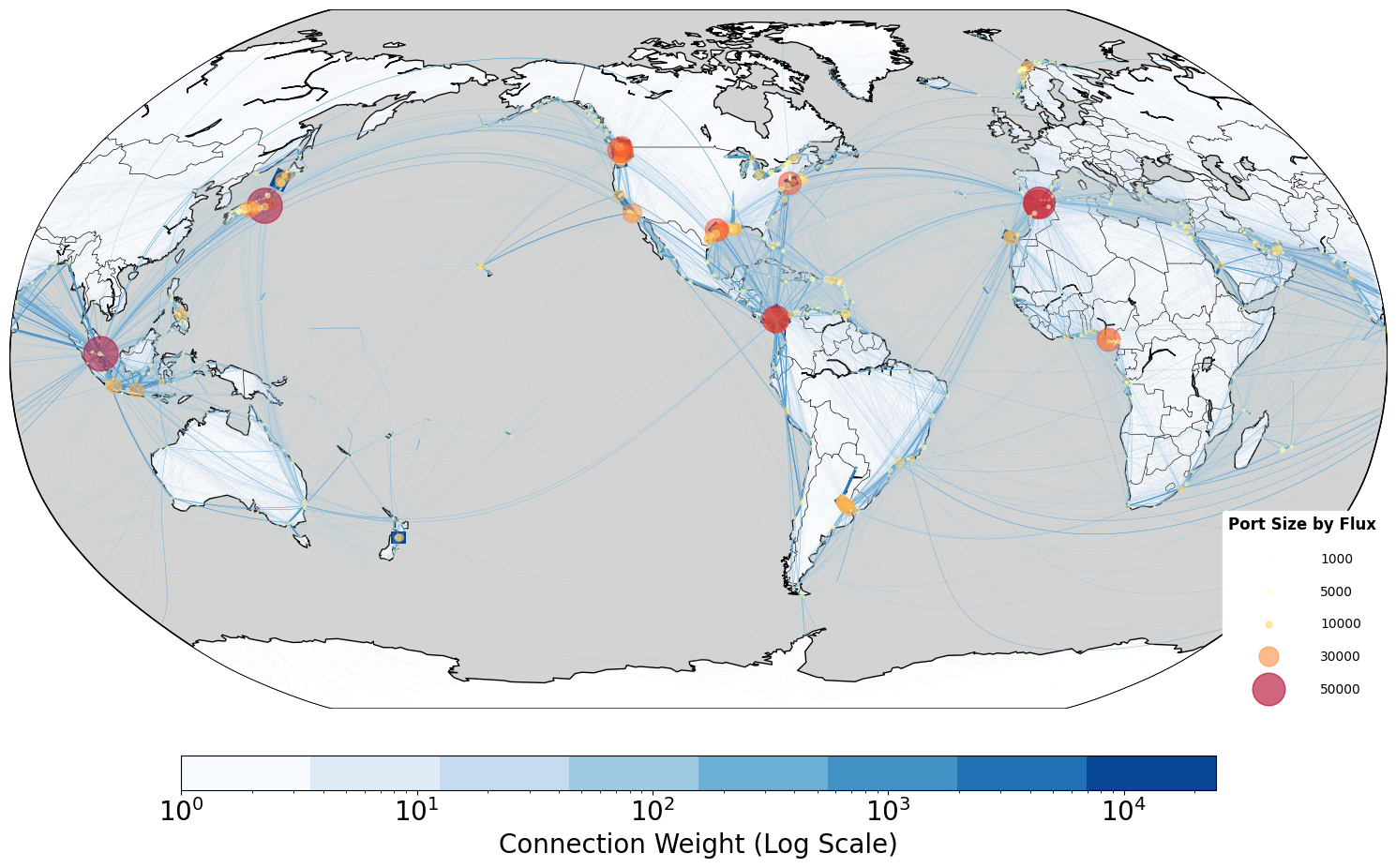} \\
    \caption{Global Shipping Network 2017--2019. Edge color and thickness are relative to the number of shipping activities per route, with darker blue and bolder lines indicating routes with higher activity levels. Port sizes in color-coded circles are scaled according to shipping fluxes to highlight their port capacity.}
    \label{fig:port_connection}
\end{figure}

We start this study by constructing a directed and weighted international shipping network based on global port visits data from 2017 to 2019, which was derived from AIS data as processed in the study of \textit{Carlini et al.}~\cite{Carlini2022UnderstandingEvolution}.
Ports and shipping connections were represented as nodes and edges in the shipping network, and the World Port Index (WPI)~\cite{WorldPortIndex} was used for port identification.
Figure~\ref{fig:port_connection} provides an overview of global shipping connections in the three years from 2017 to 2019, which shows the distribution of a total of $2,304$ ports ({\it i.e.}, nodes), $141,179$  shipping connections ({\it i.e.}, edges) and $3,573,979$ trips participating in shipping activities. 
The number of trips along the same shipping connection was used as the edge weight.

\begin{figure}[!t]
    \centering
    \includegraphics[width=\textwidth]{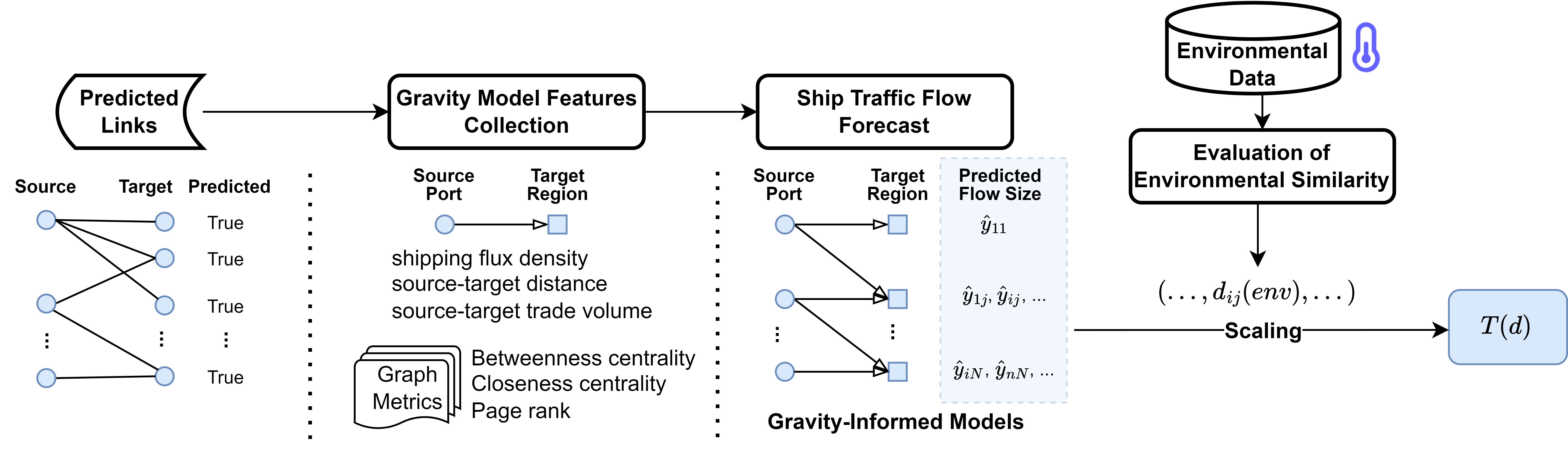} \\
    \caption{The experimental pipeline for predicting ship traffic flows with gravity-informed models, including the assessment of environmental similarity used on the ballast water risk assessment case study. The process begins with identifying and verifying predicted links between source ports and target regions. Subsequently, key features are collected, and graph metrics are extracted so they can be used to inform the gravity-based model used to forecast ship traffic flow sizes. Concurrently, environmental data pertinent to the study regions is gathered and evaluated with the aim of computing environmental similarity metrics.}
    \label{fig:ship_flow_pipeline}
\end{figure}

A detailed description of the methodological steps for creating the shipping networks, preparing the data for link prediction tasks, feature extraction, analysis, and augmentation is included and extensively discussed in the Methods section at the end of this paper.
It is noteworthy that our study was conducted during significant events ({\it e.g.}, war, global pandemic, and economic sanctions) that could have influenced the underlying shipping data.
However, all results we discuss use the same methodology, ensuring that any external factors would have had a homogeneous impact on the comparative performance of the models.

\subsection*{Network Analysis and Link Prediction}
We conducted an analysis of the shipping network to extract graph metrics. These metrics were used as part of the feature set for gravity-informed predictive models, which helped forecast traffic flow sizes on the shipping network.
We implemented link prediction as a binary classification task within the fully connected shipping network to identify the most probable shipping connections and mitigate the effect of missing links on the gravity model. These predictions are important for informing the model about the specific origin-destination (OD) pairs involved, as shown in the first two stages in Figure~\ref{fig:ship_flow_pipeline}.

\begin{figure}[!b]
     \centering
     \begin{subfigure}[b]{0.45\textwidth}
         \centering
         \includegraphics[width=\textwidth]{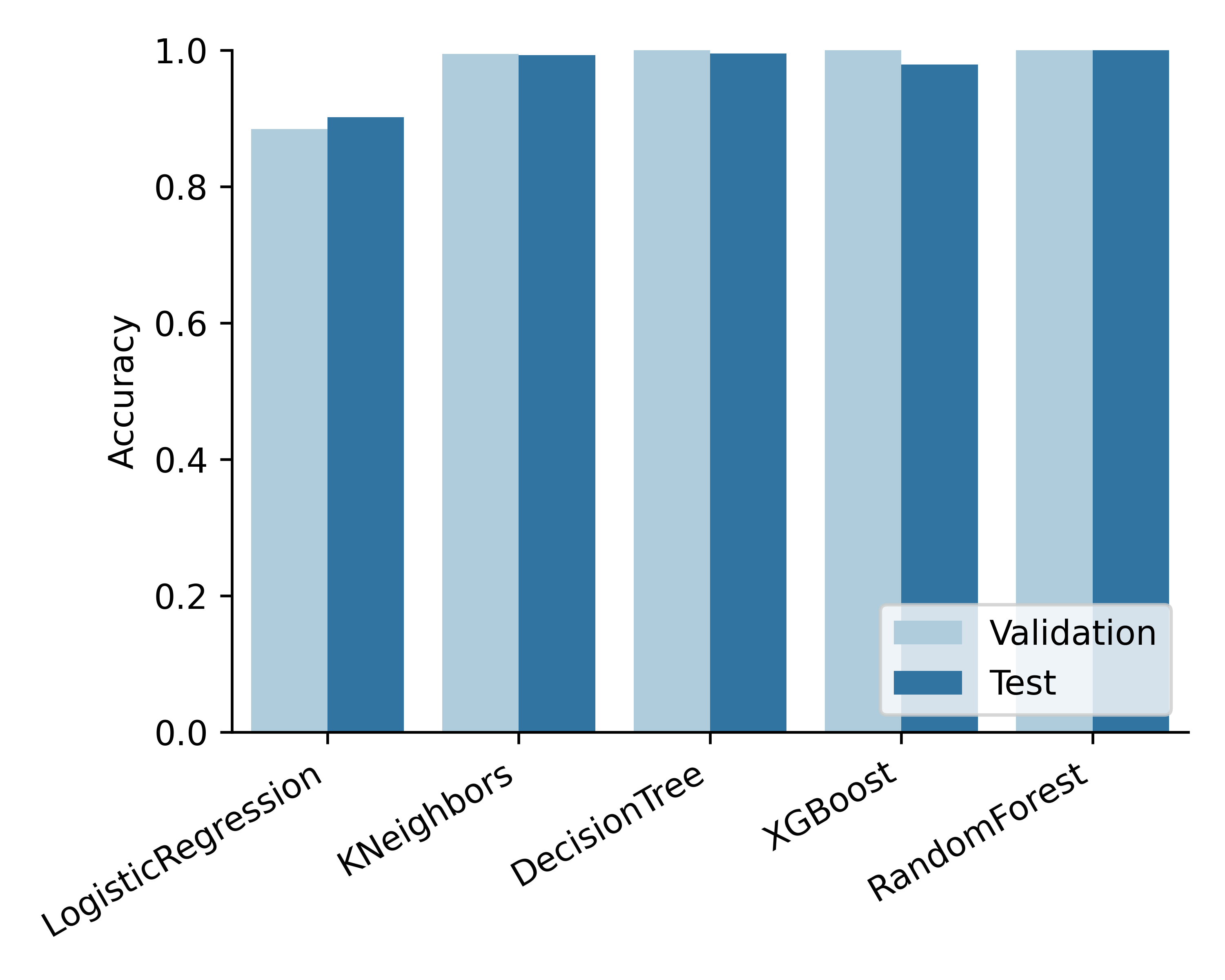}
         \caption{}
         \label{subfig:link_pred_acc}
     \end{subfigure}
     \hfill
     \begin{subfigure}[b]{0.45\textwidth}
         \centering
         \includegraphics[width=\textwidth]{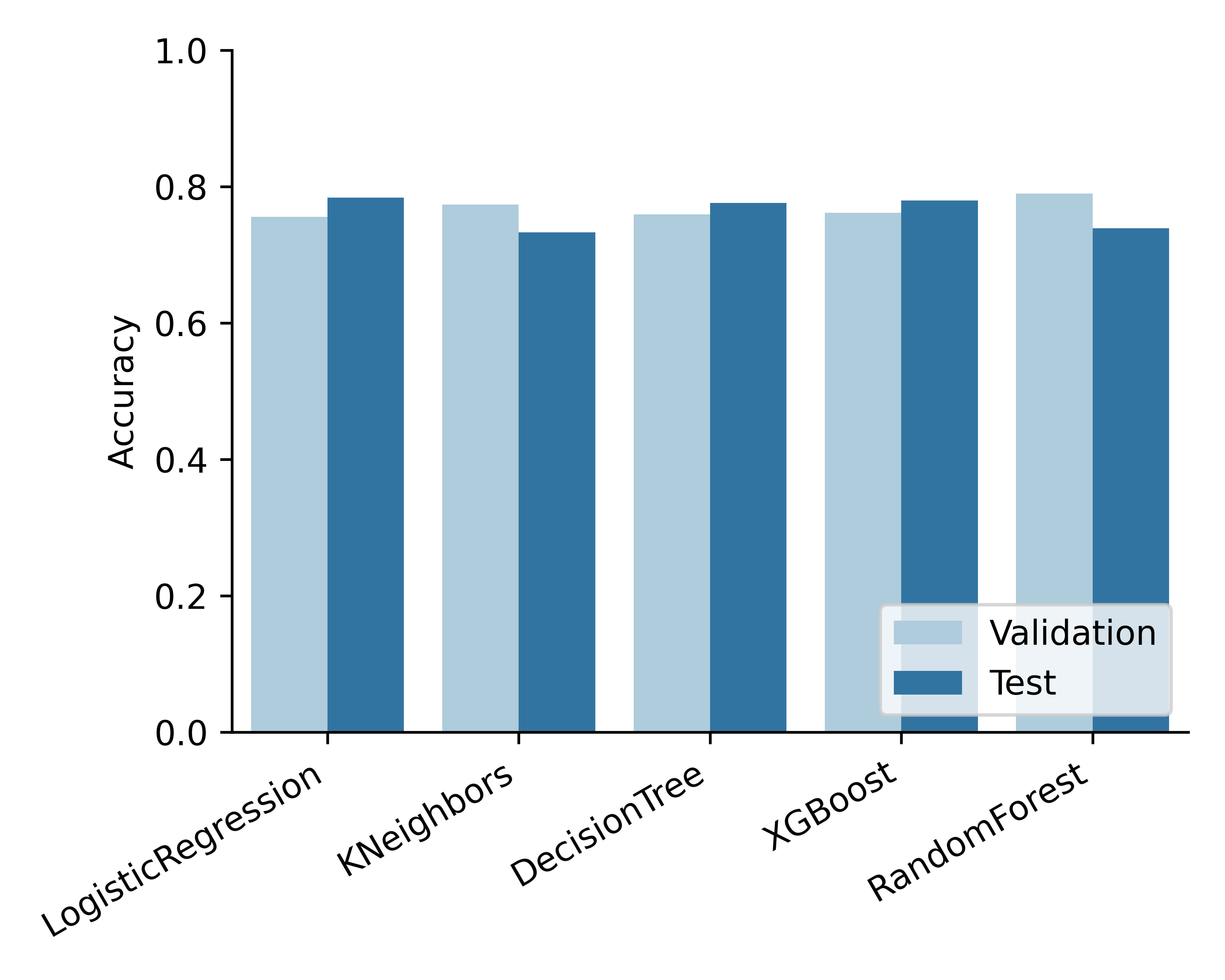}
         \caption{}
         \label{subfig:link_pred_acc_NAedgeinfo}
     \end{subfigure}
    \caption{Validation and test accuracy of classifiers in the trajectory link prediction task: \textbf{(a)} The performance of the classification task includes Haversine distance, sea route distance, and edge importance as features; and, \textbf{(b)} The classification task is carried out without the edge importance features.}
    \label{fig:link_pred_accuracy}
\end{figure}

\begin{figure}[!b]
    \centering
    \includegraphics[width=.9\textwidth]{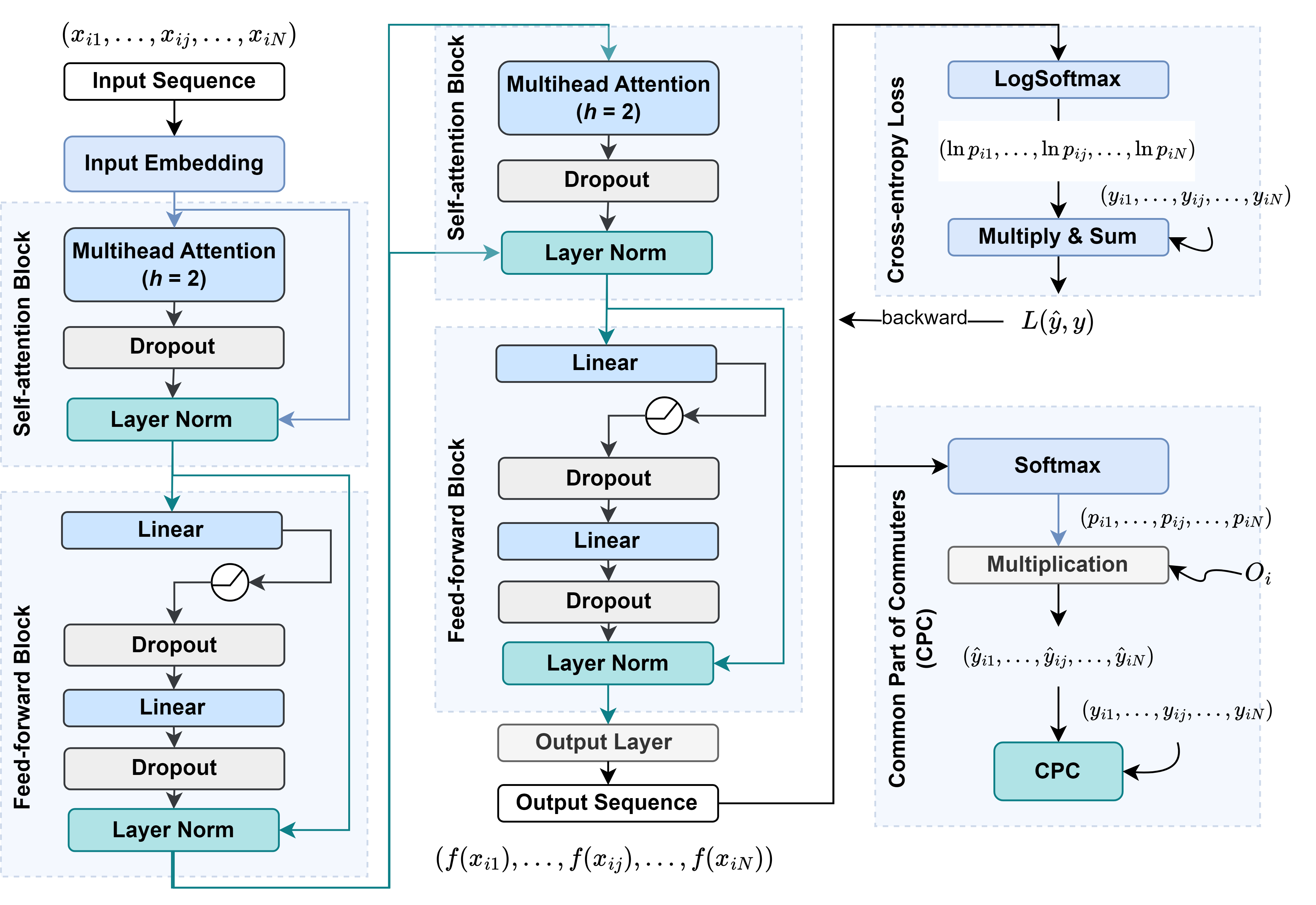} \\
    \caption{Framework of the Transformer Gravity model. The process starts with input sequences that are embedded and passed through self-attention blocks with multi-head attention, dropout, and layer normalization. This is followed by feed-forward blocks containing linear layers and dropout, resulting in the output sequence. The cross-entropy loss with log-softmax is used for training, while the Common Part of Commuters (CPC) is used for evaluation by incorporating commuting patterns $O_i$ from the input data.}
    \label{fig:model_pipeline}
\end{figure}

Figure~\ref{subfig:link_pred_acc} displays the validation and test accuracy across different classification models for the binary classification-based link prediction task.
It can be observed that apart from Logistic Regression at 89\%, all other models achieved high accuracy (above 98\%) on both the validation and the test datasets.
We further investigated the discrepancies between the model with the highest performance --- Random Forest at 100\% --- and the model with the lowest accuracy  --- Logistic Regression at 89\%.
The result revealed that the links primarily missed by Logistic Regression were those with only 1 or 2 trips and, to a lesser extent, 3-trip links.
We also observed that the high accuracy of some models could be attributed to the inclusion of normalized edge weights $w_{ij}$ ({\it i.e.}, the normalized number of trips) in the fully connected shipping network $G'$ in the edge importance calculation as per Equation~\ref{eq:edge_importance}.
Consequently, we ran predictions without edge importance as a feature to mitigate this effect for comparison.
Figure~\ref{subfig:link_pred_acc_NAedgeinfo} presents the validation and test accuracy when relying solely on Haversine and sea route distance as features.
The accuracy of these models ranges from 75\% to 79\%.
Notably, the models' performance decreases by 16\% to 20\% compared to models that use edge importance.
Retaining critical details from fluxes directly proportional to the link's existence is the key function of the edge importance feature, but creating a self-contained inference model involves more than just edge importance.
The challenge lies in selecting a model that can incorporate that information without overfitting the data.

After analyzing the results from the model shown in Figure~\ref{subfig:link_pred_acc}, we observed that the Random Forest model has a higher accuracy.
However, the Logistic Regression model acts as a filter by removing some 1- and 2-trip trajectories before predicting ship traffic flow.
These trajectories are considered unstable and unreliable for predicting future mobility flow and may be outliers.
Therefore, the Logistic Regression model identifies and removes these trajectories, leaving behind only those predicted to have more stable traffic flow between origin-destination pairs.
These trajectories are then utilized to collect features of the gravity-informed model for improving ship traffic flow prediction, such as seen in Figure~\ref{fig:ship_flow_pipeline}.
An illustrative data-flow representation of the Transformer Gravity model is depicted in Figure~\ref{fig:model_pipeline}, while a comprehensive description and formalization of the model are thoroughly given within the Methods section.

\subsection*{Performance Evaluation of Ship Traffic Intensity}
In this experiment, we aimed to evaluate and compare the performance of different models in predicting global ship traffic intensity. 
Our primary goal was to assess the effectiveness of our proposed Transformer Gravity model when compared with the original Deep Gravity model and various machine learning regression models. Through this comparative analysis, we aimed to identify the model(s) that best approximate real-world shipping mobility patterns and demonstrate superior predictive power.
In deep learning models, the number of layers refers to the depth of the neural network. Each layer in the network performs operations on the input data, gradually transforming it to make predictions. Adding more layers allows the model to learn more complex patterns and relationships within the data. Therefore, the complexity and depth of the model architecture should be adjusted to see how it affects its performance in capturing and understanding the data.
We used the proposed Transformer Gravity model with 1, 3, and 5 Transformer layers during the experimentation.
This study compared the original 15-layer Deep Gravity model with a multi-layered variant of 3, 9, and 12 layers.
Additionally, we experimented with machine learning regression models, such as linear-based, tree-based, and boosting-based models, as detailed in Table~\ref{tab:evaluation_result}.
The data from 2017 and 2018 formed the training data, while the 2019 data was used for testing.

Table~\ref{tab:evaluation_result} shows the mean, maximum, and minimum values of the models' $CPC$ across the five validation folds.
Our proposed Transformer Gravity, particularly the ones with 3 and 5-layer configurations, have achieved the best performance in cross-validation with a mean $CPC$ of 0.864, which marks a 10.5\% improvement over the top-performing 3-layer Deep Gravity variant ($CPC\ of\ 0.782$), and a 13.2\% improvement compared to original Deep Gravity model ($CPC\ of\ 0.763$) and over 49.7\% to other machine learning models, whose mean $CPC$s ranged from 0.474 to 0.577.
Meanwhile, we noticed that compared with the original Deep Gravity model, its shallower-layered variants show a better performance, as evidenced by metrics in both validation and testing.
However, none of these Deep Gravity variants can exceed the lowest mean CPC performance ($CPC\ of\ 0.846$ using a single layer) of the Transformer Gravity model.
The gap between Transformer Gravity's performance and other models indicates that our model's predicted shipping flows have a larger resemblance to the real shipping flows, highlighting our model's ability to reflect the real shipping mobility flows.
This high similarity is also reflected in the other two metrics, $NRMSE$ and $Corr$.
The 3-layer Transformer Gravity model achieved the lowest error rate ($NRMSE\ of\ 0.080$) and the highest correlation ($Corr.\ of\ 0.977$) with actual shipping flows.
These results suggest the Transformer Gravity model's superior performance in predicting global ship traffic flows over the Deep Gravity and traditional machine learning models.

\begin{table}[!ht]
    \centering
    \scalebox{0.85}{
    \begin{tabular}{r|c|ccccc|ccc|l}
    \toprule
    \multirow{2}{*}{\textit{Model Name}} &
        \multirow{2}{*}{\textit{\small Layers}} &
        \multicolumn{5}{c|}{\textit{5-Fold Cross-Validation}} &
        \multicolumn{3}{c|}{\textit{Testing}} &
        \multirow{2}{*}{\textit{\small Parameters}} \\
        \cline{3-10} & &
        \multicolumn{1}{c}{\small $CPC_{mean}$} &
        \multicolumn{1}{c}{\small $CPC_{max}$} &
        \multicolumn{1}{c}{\small $CPC_{min}$} &
        \multicolumn{1}{c}{\small $NRMSE$} &
        \multicolumn{1}{c|}{\small $Corr.$} &
        \multicolumn{1}{c}{\small $CPC$} &
        \multicolumn{1}{c}{\small $NRMSE$} &
        \multicolumn{1}{c|}{\small $Corr.$} & \\ 
        \midrule
        Linear Regression & --- & 0.474 & 0.564 & 0.353 & 0.327 & 0.457 & 0.570 & 0.327 & 0.563 & --- \\
        \hline
        Decision Tree & --- & 0.518 & 0.625 & 0.438 & 0.404 & 0.370 & 0.644 & 0.398 & 0.517 & --- \\
        \hline
        Random Forest & --- & 0.577 & 0.685 & 0.504 & 0.309 & 0.545 & 0.675 & 0.285 & 0.692 & --- \\
        \hline
        Extra Tree & --- & 0.573 & 0.686 & 0.483 & 0.311 & 0.535 & 0.699 & 0.255 & 0.767 & --- \\
        \hline
        Gradient Boosting & --- & 0.557 & 0.654 & 0.490 & 0.306 & 0.555 & 0.654 & 0.295 & 0.665 & --- \\
        \hline
        XGBoost & --- & 0.552 & 0.696 & 0.463 & 0.320 & 0.521 & 0.664 & 0.282 & 0.702 & --- \\
        \hline
        LightGBM & --- & 0.574 & 0.706 & 0.482 & 0.307 & 0.554 & 0.684 & 0.279 & 0.707 & --- \\
        \hline
        CatBoost & --- & 0.559 & 0.674 & 0.472 & 0.311 & 0.544 & 0.654 & 0.288 & 0.683 & --- \\
        \specialrule{.1em}{.05em}{.05em}
        Deep Gravity~\cite{Simini2021DeepGravity} & 3 & \textbf{0.782} & \textbf{0.797} & \textbf{0.766} & \textbf{0.209} &  \textbf{0.833} & 0.787 & 0.238 & 0.802 & 52,353 \\
        \hline
        Deep Gravity~\cite{Simini2021DeepGravity} & 9 & 0.769 & 0.776 & 0.751 & 0.220 & 0.812 & 0.783 & \textbf{0.243} & 0.797 & 249,985 \\
        \hline
        Deep Gravity~\cite{Simini2021DeepGravity} & 12 & 0.767 & 0.775 & 0.756 & 0.219 & 0.814 & \textbf{0.790} & 0.239 & \textbf{0.803} & 348,801 \\
        \hline
        Deep Gravity~\cite{Simini2021DeepGravity} & 15 & 0.763 & 0.778 & 0.729 & 0.222 & 0.808 & 0.752 & 0.273 & 0.747 & 447,617 \\
        \specialrule{.1em}{.05em}{.05em}
        Transformer Gravity$^\bigstar$ & 1 & 0.846 & 0.856 & 0.829 & 0.161 & 0.898 & 0.834 & 0.200 & 0.865 & 51,212 \\
        \hline
        Transformer Gravity$^\bigstar$ & 3 & \textbf{0.864} & 0.870 & 0.852 & \textbf{0.080} & \textbf{0.977} & \textbf{0.848} & \textbf{0.187} & \textbf{0.882} & 101,644 \\
        \hline
        Transformer Gravity$^\bigstar$ & 5 & \textbf{0.864} &\textbf{ 0.87}1 & \textbf{0.856} & 0.107 & 0.953 & 0.836 & 0.208 & 0.858 & 152,076 \\
        \bottomrule
    \end{tabular}
    }
    \caption{Performance evaluation of the Transformer, Deep Gravity models and their shallower-layered variants, and other baseline models. The cross-validation results present the mean ($CPC_{mean}$), maximum ($CPC_{max}$), minimum ($CPC_{min}$), mean Normalized Root Mean Square Error ($NRMSE$), and the mean Pearson Correlation Coefficients ($Corr.$) across five folds. Test results show the performance of the trained models on the entire test dataset. The number of layers and parameters are given for deep learning models.}
    \label{tab:evaluation_result}
\end{table}

Table~\ref{tab:evaluation_result} shows that the Transformer Gravity models show a more stable performance than the competing models.
The difference between the highest and lowest $CPC$ values for the Transformer Gravity models ranges from $0.015$ to $0.027$, which is lower than other gravity-based models ($0.019 \sim 0.049$) and machine learning models ($0.164 \sim 0.233$).
This indicates that the variance in $CPC$ is lower when the flow prediction is done with the Transformer Gravity model.
The variance can be attributed to the distribution of data samples across folds, as a random seed is set for the sample assignment.
This way, the Transformer Gravity model's performance is more consistent across different folds, meaning that given historical shipping data within the same temporal period, the Transformer Gravity model is more likely to predict shipping traffic flows with stable performance, regardless of the shipping traffic from which data subsets are sampled.

On the right side of Table~\ref{tab:evaluation_result}, we show the test results obtained by models trained on all the data from 2019 at once.
The best $CPC$ was produced by the 3-layer Transformer Gravity model ($CPC\ of\ 0.848$), which indicates that the overlap ratio between predicted and actual shipping traffic flows was very high for an unseen scenario that spanned a year.
This benefits us, as it helps us obtain more accurate shipping patterns, which we can use to evaluate the shipping intensity for ballast water risks.
Moreover, our Transformer Gravity model had a Pearson Correlation Coefficient of $0.882$.
This value indicates a strong linear relationship between predicted and actual flows, demonstrating the model's performance.

Table~\ref{tab:evaluation_result} shows that several baseline models have higher test $CPC$s than their best cross-validation $CPC$s.
This difference in performance can be attributed to the distinct feature distributions in the test data compared to the cross-validation fold data.
The spatial and temporal dependencies intrinsic to the shipping data can also affect these evaluation results.
Our training set includes shipping data from 2017 and 2018, while the test set comprises 2019.
Since both datasets represent complete years, the test data is expected to be more closely aligned with the spatial distribution of the training data.
This may account for the discrepancy between the higher test performance and the observed validation results for these models.
However, the evaluation of the Transformer Gravity model shows a better result in validation sets, which follow a more conventional pattern and have minimal variance over the cross-validation folds.
This indicates that our model has gained a deeper understanding of the connections between data features and is robust enough to handle the impact of varying data distributions within the dataset.
This shows the potential of the model to make accurate predictions on unseen datasets.

Table~\ref{tab:evaluation_result} also reveals that the performance of the Transformer Gravity model varies with the addition of layers.
An increase in the number of the Transformer encoder layers initially enhances the model's performance, as evidenced by the 3-layer Transformer Gravity model results compared with the single-layer model shown in the table.
However, this performance increase is not unlimited.
Comparing the 3-layer Transformer Gravity with the 5-layer one, its performance tends to plateau as the model grows deeper with better-optimized parameters.
Therefore, while stacking Transformers can be beneficial, optimizing the number of layers is essential to maximize the model's performance.

Similarly, the Deep Gravity model also needs an optimized configuration.
The original 15-layer Deep Gravity model includes 256 hidden dimensions for layers 1 to 5 and 128 dimensions for layers 6 to 15.
Due to the relatively large performance variation among its validation folds ($CPC_{max}\ of\ 0.778$ and $CPC_{min}\ of\ 0.729$), we adjusted the number of layers and experimented with shallower versions of the model for comparison.
We followed the 1:2 ratio of the 256 and 128 dimensions in the original Deep Gravity to configure the 3, 9, and 12-layer variants.
As per Table~\ref{tab:evaluation_result}, we observed that Deep Gravity models with fewer layers delivered better performance than the original deep-layered model consisting of 15 layers.
However, the performance of Deep Gravity models decreased with the addition of more layers.
The 3-layer model of Deep Gravity performed the best, while deeper networks could not yield better results.

\subsection*{Performance Evaluation over the Risk Assessment of Invasive Species}
As part of our study, we conducted a final experiment using the models we proposed on the global shipping network.
We assessed the risk of NIS \textcolor{black}{introduction} associated with shipping flows using the BWRA decision tool employed by Transport Canada~\cite{Bradie2021DecisionSupport,etemad2022developing}.
We aimed to demonstrate the performance of our proposed model over a real application scenario, whereby by having a more accurate overview of the shipping traffic between regions, we can improve real-world risk analysis. 
In this case, by forecasting shipping intensities on maritime pathways with varying levels of invasion threats, regulatory agencies can have access to an estimation that can aid data-driven ocean regulation and policy-making, especially by identifying high-risk pathways that need enhanced monitoring and regulations to support maritime biosecurity. 
The risk model we used produces an environmental distance value that compares the dissimilarities between a ship's origin and destination. When a ship arrives at a destination and comes from an origin with similar environmental characteristics ({\it i.e.}, higher) as the destination port, it is more likely to be inspected by the responsible authorities.
The model considers environmental conditions such as sea surface temperature and salinity at the locations where ballast water is taken and discharged.
We gathered environmental variables at these locations, including minimum, maximum, annual temperature, and yearly salinity.
We compiled them into a vector called $\nu_{i} = \langle t_{(i)_{min}}, t_{(i)_{max}}, t_{(i)}, s_{(i)} \rangle$.
Through the vector $\nu_{i}$, we proceed to calculate the environmental distance using the element-wise Euclidean distance calculation:
\begin{equation}
    \hfill
    \begin{aligned}
        d_{ij} (env) &= \sqrt{\sum_{k=0}^{|\nu|}(\nu_{i_k} - \nu_{j_k})^2}
    \end{aligned}
    \hfill
    \label{eq:distance_calculation}
\end{equation}

\noindent
Based on Equation~\ref{eq:distance_calculation}, a smaller $d_{ij} (env)$ value suggests a higher environmental similarity between the origin and destination, typically meaning a higher risk of NIS invasion via ballast water and vice versa.

\textcolor{black}{To evaluate the environmental distances between the shipping source and destination}, we first integrated port data from \textit{World Port Index}~\cite{WorldPortIndex} with the environmental conditions in \textit{Global Port Environmental Data}~\cite{Bailey2020EnvironmentalData}, which includes the annual average salinity, annual average, minimum and maximum temperature nearest to each port.
\textcolor{black}{For each port involved in the 2019 shipping data, we calculated the environmental distances based on the environmental conditions at both the source and destination ports using Equation~\ref{eq:distance_calculation}.
These environmental distance values inform the threat levels associated with various shipping pathways. Then, we integrated the environmental distance values with the shipping intensity by weighting environmental distance according to the number of trips on specific shipping routes. 
This combination of both threat levels and shipping intensity enables the mapping of the overall distribution of shipping activities across the global network at varying invasion risk levels. 
To assess the dissimilarity in risk assessment results between predicted and actual shipping flows, we examined the distributions of environmental distances using the estimated trips from} the Transformer Gravity model (denoted as $T(d)_{TG}$), the Deep Gravity model ($T(d)_{DG}$), and the actual shipping flows from 2019 (represented as $T(d)_{true}$).

\begin{figure}[!ht]
    \centering
    \includegraphics[width=.95\textwidth]{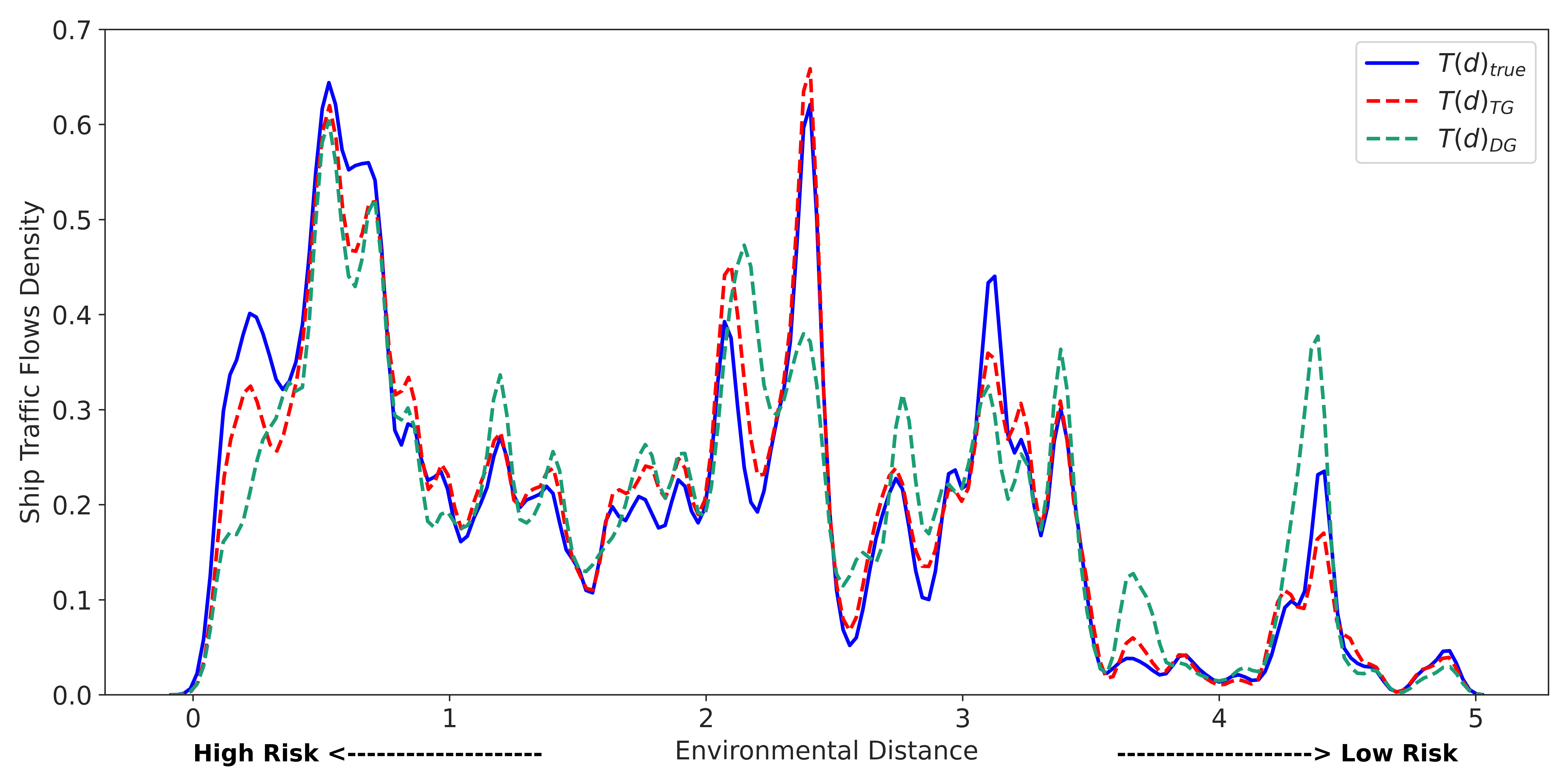} \\
    \caption{Distribution of environmental distances for shipping flows in 2019. The blue curve represents the true shipping flows, while the dashed red and green curves depict the predictions from the Transformer Gravity (TG) and Deep Gravity (DG) models, respectively. The x-axis measures the environmental distance; smaller values indicate higher risk levels, and larger values indicate lower risks.}
    \label{fig:environmental_distance}
\end{figure}

Figure~\ref{fig:environmental_distance} shows the distributions of the three distance groups: $T(d)_{true}$, $T(d)_{TG}$, and $T(d)_{DG}$, from which we can find that $T(d)_{true}$ and $T(d)_{TG}$ are closely aligned in their trends, even in minor fluctuations.
In contrast, $T(d)_{DG}$ is more differentiated from $T(d)_{true}$.
To quantify the alignment between $T(d)_{TG}$ and $T(d)_{true}$, we calculated Pearson's correlation coefficients for the two groups of environmental distances, reaching the value of $0.889$.
This high coefficient indicates a strong linear correlation between the environmental distances associated with actual and predicted shipping flows.
The figure also reveals that the consistency between $T(d)_{TG}$ and $T(d)_{true}$ is more pronounced where the environmental distance is greater, implying a lower risk of invasion.
Conversely, the two groups show discrepancies \textcolor{black}{in the higher invasion risk interval (i.e., smaller environmental distances), which can be attributed to the differences between the predicted and the actual shipping flows. These discrepancies can be influenced by an important feature, {\it i.e.}, the bilateral trading volume, which quantifies the annual trading volume between two countries in US dollars.
For inner-country shipping activities, the bilateral trading volume} becomes zero and no longer positively associates trade with shipping flows. \textcolor{black}{These inner-country routes} are usually high-risk due to the environmental similarity between origins and destinations.
Further, there may also be critical factors in predicting unexplored high-risk shipping connections, suggesting potential for further studies.

\section*{Discussion}

The Transformer Gravity model is built based on a stack of multi-head attention blocks.
It incorporates features from the gravity-informed model, including the shipping fluxes, origin and destination locations, and the geographical distance between them.
We enhanced the feature set by integrating extra factors, such as bilateral trade information and graph metrics from the global shipping network, to improve the model's ability to predict ship traffic flows.
We validated our approach by comparing it with established gravity models and machine-learning regression techniques to showcase its advancements.

Our findings established the superiority of the Transformer Gravity model across several performance indicators.
Notably, the model achieved the highest CPC with minimal variance across different data folds, indicating its robustness and adaptability to new datasets.
The best average CPC in the cross-validation set was 0.864 for the 3 and 5-layer models.
Besides, the 3-layer one demonstrated the lowest mean error, 0.080, and the highest mean correlation, 0.977.
Variations in performance among baseline models indicated their inadequacies in dealing with diverse data scenarios, thus reinforcing our model's resilience.

The high performance of our Transformer Gravity model for ship traffic flow prediction contributes to a more accurate representation of shipping intensity in the risk assessment of invasive species through BWRA.
Despite some discrepancies in the high-risk interval, the evaluation results show an overall low error of $0.208$ and a high similarity between predicted and actual environmental distances among port pairs scaled by ship traffic flows.
This has contributed to a more accurate risk assessment for the spread of NIS through ballast water and thus provides a valuable reference for future global ballast water risk management.
Specifically, the current Ballast Water Management Convention of the International Maritime Organization (IMO) regulates ship ballast water exchanges, mandating they occur away from coastal areas and at specified depths to decrease the survival and spread of invasive species~\cite{IMO-BWMC}.
\textcolor{black}{To further reduce the survival of NIS, various ballast water treatment methods have been implemented. These include ballast water exchange and on-board ballast water treatment systems (BWTSs) that employ physical, mechanical, and chemical approaches. However, none of these methods are effective at all times~\cite{Tsolaki2009BallastWaterTreatment}. Although IMO regulations became mandatory in 2017, a regional study indicates that as of 2019, less than 20\% of ships arriving in the United States and Australia were equipped with BWTSs~\cite{Gerhard2019InstallationUse}. This gap in compliance and the partial effectiveness of current ballast water treatments highlight the necessity for advanced predictive models.}
The Transformer Gravity model plays a pivotal role in finding ports characterized by intense shipping activity through its ability to forecast shipping intensity across diverse maritime pathways. 
This predictive capability is paramount, especially for routes forecasted to exhibit elevated shipping volumes and a heightened risk of NIS introduction. By identifying these high-risk routes, our model serves as a valuable tool for governmental agencies, urging them to strengthen monitoring and management efforts. 
This approach is essential to uphold compliance with IMO regulations and mitigate the risks of NIS.

In summary, the Transformer Gravity model proposed in this study has markedly improved the performance of ship traffic flow forecast, outperforming other gravity-informed models.
It has also shown high-performance consistency when applied to different data subsets with various spatio-temporal distributions.
Future mobility studies and applications can enhance the model's explainability and transparency by exploring the intrinsic relationships between various features and the flow prediction results.
The exploration of such aspects can improve the model's interpretability and help the model's application in domains where a high degree of accuracy is expected on the OD flow representation.

\section*{Methods}
\setcurrentname{Methods}
\label{sec:methods}

\subsection*{Global Shipping Network Representation}
The global shipping network can be described as $G=(V, E, W)$, where $V$ represents the set of ports, $E$ is the set of shipping routes connecting pairs of ports, and $W$ is the collection of edge weights.
In this context, each weight in $W$ corresponds to the number of individual trips $T_{ij}$ between ports $i$ and $j$:
\begin{equation}
    \hfill
    W = \{ w_{ij} : w_{ij} = \sum_{t} T_{ij}(t) \, \forall (i, j) \in E \}
    \label{eq:edge_weight}
    \hfill
\end{equation}

\begin{figure}[!ht]
    \centering
    \includegraphics[width=.95\textwidth]{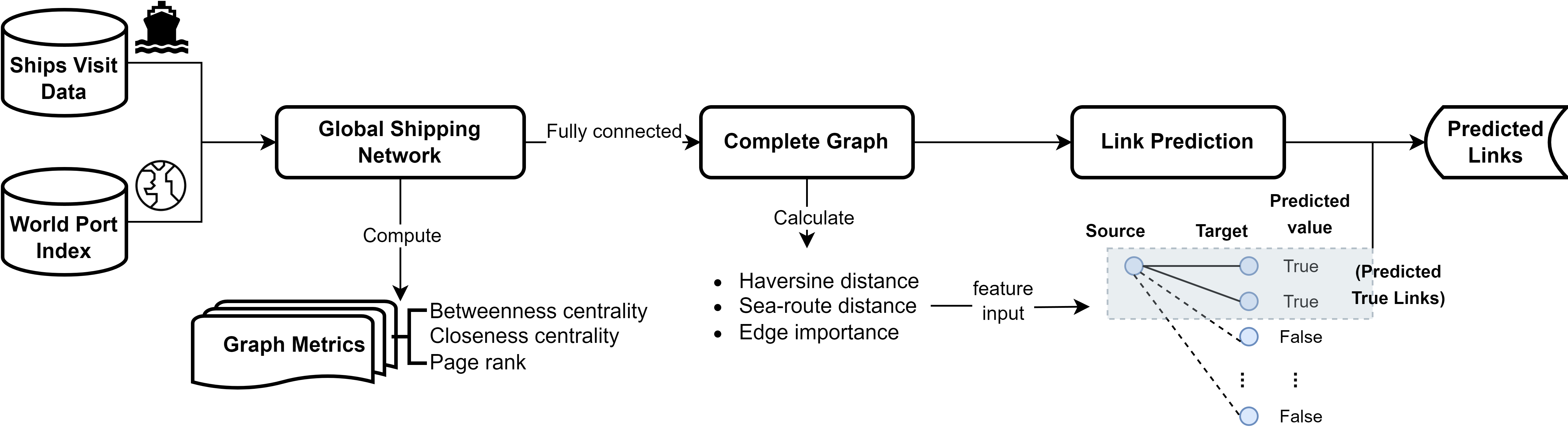} \\
    \caption{The experiment pipeline for analyzing and predicting links in the global shipping network from 2017 to 2019 includes constructing the global shipping network, calculating centrality and PageRank graph metrics, fully connecting the shipping network for disconnected and weakly connected components, and performing link prediction to identify potential real shipping connections.}
    \label{fig:link_pred_pipeline}
\end{figure}

With the global shipping network defined, we performed network analysis to extract graph metrics as features for our proposed gravity-informed predictive model.
We also conducted link prediction to identify potential origin-destination (OD) pairs within current shipping traffic, thus providing pre-knowledge to the predictive models.
Figure~\ref{fig:link_pred_pipeline} illustrates the pipeline of shipping network analysis and the link prediction process, which precedes the gravity-informed predictive models forecasting the ship traffic flows.

\subsection{Complex Network Metrics Computation}
We analyzed the shipping network by calculating various graph metrics for each node.
These metrics included betweenness centrality, closeness centrality, and PageRank.
The weights of the edges utilized in these metrics computations correspond to the number of trips along these individual shipping routes, as defined in Equation~\ref{eq:edge_weight}.
Subsequently, we introduced the graph metrics used for shipping network analysis:

\noindent\textit{\textbf{Betweenness centrality $(C_B)$:}}
The Betweenness identifies how often a particular vertex is present in the shortest paths of a graph.
A higher metric value indicates that the vertex is present in a larger number of shortest paths.
In Equation~\ref{eq:betweenness_centrality}, $\sigma(u,v)$ represents the number of shortest paths between vertices $u$ and $v$, and $\sigma(u,v|i)$ represents the number of those paths that pass through vertex $i$.
The \textit{Edge-Betweenness}, which assigns a Betweenness value to each edge in a graph, is worth noting.
This variant considers $\sigma(u,v|i)$ as the shortest paths between vertices $u$ and $v$ that pass through edge $i$.
In the shipping network, betweenness centrality quantifies the frequency with which a port serves as an intermediary on the shortest paths between other ports; ports with high betweenness behave as bridges in the shipping network.
\begin{equation}
    \hfill
    {C_B(i)} = \sum_{\left<u, v\right> \in V}{} \frac{\sigma(u,v|i)}{\sigma(u,v)}, ~ u \neq v
    \label{eq:betweenness_centrality}
    \hfill
\end{equation}

\noindent\textit{\textbf{Closeness centrality $(C_C)$:}}
The Closeness measures how closely a vertex is connected to all the other vertices in a network, as determined by the shortest paths between them (refer to Equation~\ref{eq:closeness_centrality}).
A vertex with higher centrality is considered more central and has a shorter average distance to all other vertices in the network~\cite{Kas2013:IncrementalCloseness}.
In a port network, the closeness-central ports are easily accessible from all other ports in the shipping network. They are expected to have higher vessel traffic, providing insight into trading behavior and economic relationships inherent to their country and cities.
\begin{equation}
    \hfill
    C_C(i) =  \frac{|V|-1}{\sum_{j=1}^{|V|} d_{ij}^{N}}, ~ i \neq j
    \label{eq:closeness_centrality}
    \hfill
\end{equation}

\noindent\textit{\textbf{PageRank $(P_{ij})$:}}
The PageRank~\cite{Page1999PageRankCitation} is based on a mathematical model known as the stochastic Markovian process.
This model defines a probability distribution over a set of states, where the probability of transitioning from one state to another is solely correlated with the state immediately preceding it:
\begin{equation}
    \hfill
    P_{ij} = \mathbf{P}(X_{n+1}=j\mid X_n=i)
    \label{eq:markovian-process}
    \hfill
\end{equation}
The algorithm assesses the significance of a vertex concerning the significance of the vertices linked to it.
This way, it measures the vertex contribution based on the number of outgoing edges each adjacent vertex has, ensuring the uniqueness of the edge.
The algorithm calculates the stationary transition probability matrix to determine vertex importance.
The values obtained from this calculation indicate the significance of the vertices based on their access probability.
Equation~\ref{eq:stationary-matrix} illustrates the stationary matrix, where $\pi^{(n)}$ denotes the probability matrix at time $n$, and $\mathbf{P}$ is the transition probability matrix.
\begin{equation}
    \hfill
    \pi^{(n-1)}\mathbf{P} = \pi^{(n)}
    \label{eq:stationary-matrix}
    \hfill
\end{equation}
As PageRank identifies essential nodes in a graph ---
ports with high metric values are more influential and likely to be frequently visited by ships from other important ports.

After calculating three different graph metrics, we noticed that they all showed positive non-linear correlations with each other.
This means that as one metric increased, the others also tended to increase.
Although each metric has a distinct interpretation, we found five ports from different bodies of water that had high values in all three metrics (see Table~\ref{tab:graph_metric}).
These ports are particularly noteworthy because they excel in multiple areas.
Many of the most influential ports are located at crucial marine traffic junctions that connect different oceans, such as the Gulf of Suez, the Gulf of Panama, and the Strait of Malacca.

\begin{table}[!ht]
    \centering
    \scalebox{0.98}{
        \begin{tabularx}{\textwidth}{c|ccclll} \hline
            Port Name~\cite{WorldPortIndex} & $C_B(i)$ & $C_C(i)$ & $P_{ij}$ & Water Body & Country & Region \\ \hline
            Keppel & 0.021 & 0.562 & 0.012 & Strait of Malacca & Singapore & South-eastern Asia \\
            Europa Point & 0.018 & 0.553 & 0.012 & Strait of Gibraltar & Gibraltar & Southern Europe \\
            Puerto Cristobal & 0.018 & 0.539 & 0.007 & Caribbean Sea & Panama & Latin America \& Caribbean \\
            As Suways & 0.018 & 0.532 & 0.003 & Gulf of Suez & Egypt & Northern Africa \\
            Balboa & 0.018 & 0.523 & 0.006 & Gulf of Panama & Panama & Latin America \& Caribbean \\ \hline
        \end{tabularx}
    }
    \caption{Ports with the highest graph metric values in betweenness centrality ($C_B(i)$), closeness centrality ($C_C(i)$), and PageRank ($P_{ij}$) with port information~\cite{WorldPortIndex}, including water body, country, and region.}
    \label{tab:graph_metric}
\end{table}

\textcolor{black}{Given that betweenness and closeness centrality measures involve the calculation of shortest distances, we also applied the reciprocals of the weights defined in Equation~\ref{eq:edge_weight} to calculate the graph metrics. This transformation aligns with interpreting weights as distances. However, the centrality rankings of the ports listed in Table~\ref{tab:graph_metric} remained largely unchanged even after applying the reciprocal weights. This suggests that the topology and connectivity patterns are more critical attributes in the real-world shipping network characterized by sparsity and a highly skewed degree distribution. Furthermore, our shipping network modeling strategy aligns with the principles of gravity models. The strength of relationships is determined by the masses of sources and destinations, with larger objects driving more gravitational pull.}

\subsection*{Fully Connected Shipping Network}
A component is a group of vertices connected to each other, and a network with more than one group of vertices that are not connected is called a non-connected graph.
In an arbitrary graph, vertices $i$ and $j$ are in the same component $G'=\{V',E'\}$, $V' \subseteq V$, if and only if $\{\forall i \in V', \forall j \in V'| d_{ij} < \infty\}$, meaning that it is possible to travel from any vertex $i$ to any vertex $j$ in a finite number of steps, where $d_{ij}:V \times V \rightarrow \mathbb{R}$ is a function that returns the distance between any two vertices~\cite{cormen2001introduction}.
Connected components for directed graphs are divided into weakly and strongly connected components, both of which identify absorbing regions.

When analyzing the shipping network, we observed the presence of 2 disconnected components and 13 components weakly connected ({\it i.e.}, these subgraphs are unilaterally connected).
These disconnected and weakly connected segments pose a challenge for our proposed framework, which relies on integrating features across all possible destinations from each source port.
Such disconnections can compromise the model's ability to generate accurate or well-defined predictions for shipping flows between isolated areas.
Thus, we transformed the original network into a complete graph by connecting each pair of nodes $V_i$ and $V_j$ in the shipping network $G$. We assigned a small weight, $w'=0.1$, to these newly established links and denoted the complete graph as $G'$.
More specifically, we set the weights of these non-existent edges to be inversely proportional to the distance. This approach considers the low probability of vessels leaving one port from a component and directly reaching another, ensuring an accurate representation of the network's connectivity.
This action mitigates the issues arising from data sparsity and establishes a uniform data structure, thereby enhancing the robustness of the flow estimation framework.
With the fully connected shipping network $G'$, we proceed with the flow estimation framework by forecasting whether a trajectory exists by solving a link prediction problem in $G'$, which can be framed as a binary classification task within machine learning models.
This action provides concrete source-destination pairs well-prepared for building feature sets and modeling the gravity force.

\subsection*{Network Feature Extraction, Representation, and Purpose}
To perform link prediction, we first separated the shipping data from 2019 for testing and retained the data from 2017 and 2018 for training and validation.
Then, we prepared features for this binary classification task.
We calculated the Haversine distances between every pair of ports, which is defined in Equation~\ref{eq:geodesic_distance}, measures the shortest distance between two vertices $i \in V$ and $j \in V$ on a sphere's surface~\cite{Konstantopoulos2012:ProjectiveGeometry}.
This metric incorporates the latitudes $\phi_{i}$ and $\phi_{j}$ of vertices $i$ and $j$, the difference $\bigtriangleup^{\phi}_{ij}$ between $\phi_{i}$ and $\phi_{j}$, the difference $\bigtriangleup^{\lambda}_{ij}$ between their longitudes $\lambda_{i}$ and $\lambda_{j}$, and the Earth's radius $\mathbf{R}$ (6,371 km) --- all values are in radians.
\begin{equation}
    \hfill
    d_{ij}^{E} =  2 \times arcsin\left( 
    sin^2\left({\bigtriangleup^{\phi}_{ij}}\right) + cos\left(\phi_{i}\right) \times cos\left(\phi_{j}\right) \times sin^2\left({\bigtriangleup^{\lambda}_{ij}}\right)
    \right) \times \mathbf{R}
    \label{eq:geodesic_distance}
    \hfill
\end{equation}

However, Haversine distances only provide the geodesic approximation and cannot capture the real sea routes that the ships have traveled.
Therefore, we also computed the sea-route distances between port pairs.
We obtained a more accurate representation using a Python package that models the shortest routes and calculates the sea route distances using historical trajectories~\cite{GentHaliliSearoutepyPython}.
Finally, we used these distances to evaluate the importance level $I_{ij}$ for each edge $\langle i, j \rangle$ in the complete graph $G'$.
Inspired by straightness centrality measuring the node connectivity by the straightness of the shortest distance~\cite{Crucitti2006Centrality}, this metric combines the normalized flow size $\hat{w}_{ij}$ and the normalized Haversine distance $\hat{d_{ij}^E}$, using a small constant $\epsilon$ to prevent division by zero.
Connections with high shipping flows and shorter distances are deemed more important:
\begin{equation}
    \hfill
    I_{ij} = \frac{\hat{w}_{ij}}{\hat{d_{ij}^E}+\epsilon},~
    i, j \in V', ~i \neq j
    \label{eq:edge_importance}
    \hfill
\end{equation}
Following is an explanation of the shortest distance and straightness centrality concepts in network analysis, from which we drew inspiration for our edge importance metric (Equation~\ref{eq:edge_importance}):

\noindent\textit{\textbf{Shortest distance $(d_{ij}^{N})$:}}
A path $S$ between two vertices $i$ and $j$ is a sequence of connected vertices $S^n = \left<v_1, v_2, \text{...}, v_{q-1}, v_q\right>$, where each consecutive vertex is connected through an edge $\left<S^n_m, S^n_{m+1}\right> \in E$ for all $m \in [1,|S^n|[$.
The shortest directed path $d_{ij}^{N}$ is obtained by minimizing the weight function ${\mathit{f}:E^n \rightarrow \mathbb{R}}$, which describes the cost of the paths among all possible paths $\mathbb{S} = \{S^{1}, S^{2}, \ldots, S^{n}\}$ between vertices $i$ and $j$~\cite{VanSteen2010:GraphTheory}.
The goal is to find the path with the minimum cost, which is determined by the sum of the weights of the edges.
The weight is the straight-line distance between the vertices, such as in Equation~\ref{eq:shortest-path}.
\begin{equation}
    \hfill
    d_{ij}^{N}= min\left(\sum_{m=1}^{|S|-1} \mathit{f}(\left<S^n_m, S^n_{m+1}\right>), \forall S^n \in \mathbb{S} \right)
    \label{eq:shortest-path}
    \hfill
\end{equation}

\noindent\textit{\textbf{Straightness centrality $(C_S)$:}}
This metric measures the straightness of paths connecting vertices $i$ and $j$.
It does so by comparing the deviation of the geodesic distance $d_{ij}^{E}$ and the shortest path distance $d_{ij}^{N}$ that links them~\cite{Vragovic2005:StraightnessCentrality}.
A high centrality value indicates the existence of connections with distances close to the geodesic one.
When the two distances match, it is the optimal scenario for communication between vertices.
\begin{equation}
    \hfill
    C_S(i) = \frac{1}{|V|-1}\sum_{j=1}^{|V|} \frac{d_{ij}^{E}}{d_{ij}^{N}},~i \neq j
    \label{eq:straightness_centrality}
    \hfill
\end{equation}

Next, we incorporated the Haversine distance (Equation~\ref{eq:geodesic_distance}), sea route distance~\cite{GentHaliliSearoutepyPython}, and edge importance (Equation~\ref{eq:edge_importance}) into the feature set. We then employed machine learning classification models, including Logistic Regression, KNeighbors, Decision Tree, XGBoost, and Random Forest, to determine the existence of a link between two ports.
Upon labeling the real (true class) and pseudo (false class) links in the complete network $G'$, we observed a significant class imbalance (2.3\% real links and 97.7\% pseudo links); consequently, we employed stratified sampling of the pseudo links based on their spatial distribution to balance the number of examples in each class in a manner that preserved the data's characteristics.
The models were engaged in a binary classification task, where we utilized 75\% of data from 2017 and 2018 for training and 25\% for validation.
During the training phase, a 5-fold grid search cross-validation was implemented to fine-tune the hyperparameters of each model.
Finally, we tested the models on unseen data from 2019, reinforcing the validity of our approach.

\subsection*{Features used with Gravity-Informed Models}

Based on predicted shipping trajectories, we have origin-destination pairs for preparing gravity model features.
Features incorporated in gravity-informed models are listed in Table~\ref{tab:feature_info}, including shipping fluxes, distances, international bilateral trade volume~\cite{Harvard2019InternationalTrade}, and the centrality graph metrics extracted from the global shipping network. 
In particular, the bilateral trade volume data is country-based. Therefore, we used the source and destination countries to extract trade volume data and integrate it into the feature set.

\begin{table}[!ht]
    \centering
    \scalebox{0.98}{
    \begin{tabularx}{\textwidth}{l|X}
    \hline
        \textbf{Feature name} & \textbf{Feature information} \\
        \hline
        \textit{Original port fluxes} & Shipping fluxes at source port \\
        \hline
        \textit{Destination region fluxes} & Total shipping fluxes at the destination geographical region \\
        \hline
        \textit{Distance} & Geodesic distance between the source port to the destination region center \\
        \hline
        \textit{Bilateral trading volume} & Exportation volume in US dollars from source country to destination country\\
        \hline
        \textit{Betweenness centrality} & Betweenness centrality at the original port and the median betweenness centrality in the destination region\\
        \hline
        \textit{Closeness centrality} & Closeness centrality at the original port and the median closeness centrality in the destination region \\
        \hline
        \textit{Page rank} & Page rank at the original port and the median PageRank in the destination region \\
        \hline
    \end{tabularx}
    }
    \caption{Detailed information of features used in Transformer Gravity and other gravity-informed models.}
    \label{tab:feature_info}
\end{table}

\subsection*{Transformer Gravity Model}

In this study, we incorporate the self-attention mechanism of the Transformer architecture~\cite{Vaswani2017AttentionAll} into our proposed framework.
Compared to the conventional MLPs structure, the self-attention mechanism can inspect the input sequence and weigh to identify and prioritize the most relevant elements for generating the output. This characteristic enables our model to capture crucial dependencies in vessel mobility effectively flows.
Additionally, the self-attention mechanism accomplishes high performance with fewer parameters, making it a computationally efficient model.
This section shows how we model the \textit{Transformer Gravity}, combining the characteristics of the Gravity Model and the self-attention mechanism.

\subsubsection*{Problem Definition}

Based on the pairs of source-destination ports obtained from the previous link prediction step, we encode the destination ports into 17 geographical regions according to the ISO-3166 standard~\cite{ISO3166}, where ports within regions are expected to have a similar set of organisms and, therefore, share similar habitat.
Given the limitations in the granularity of our data, which does not extend to cover all geographic areas, this regional categorization makes it feasible to model and incorporate all intra-region locations within the landscape.
Over the encoded representation of regions, we now aim to estimate the sizes of shipping mobility flows between each source-destination pair $(P_i, P_j)$, where $P_i$ is the source port and $P_j$ is the destination port that pertains to a unique geographical region.
A ship departing from a source port may have one or more destination ports in the same or different regions, and following the Deep Gravity~\cite{Simini2021DeepGravity} method, the goal is to estimate probabilities of the ships traveling to these geographical regions, becoming a multiclass classification task.

\noindent\textit{\textbf{Predict target:}}
We compile a set of 10 features from various sources for each pair $(P_i, P_j)$, such as shipping fluxes at ports, geodesic distances between source and destination, bilateral trade volume source from \textit{The Atlas of Economic Complexity}~\cite{Harvard2019InternationalTrade}, the graph metrics computed with the global shipping network; detailed information about these extracted features is provided in Table~\ref{tab:feature_info} in the Methods.
We represent the feature vector for each source-destination pair as $x_{ij} = \left<m_1, m_2, \ldots, m_{10}\right>$.
Given the ships from each source port travel to multiple destination regions, these feature vectors are aggregated into a single data sample $X_i = \{x_{i1}, x_{i2}, \ldots, x_{iN}\}$, where $N$ is the number of destination regions in the data sample, and $1 \leq N \leq 17$.
Each destination region is represented as a class, so the prediction has $N$ classes for a sample.
Since samples of varying lengths cannot be wrapped to a tensor for batch processing, we set the batch size to 1 for model input.
Using $\hat{y}_{ij}$ to represent the estimated size of the mobility flow between $(P_i, P_j)$, which is the target of the prediction, we have:
\begin{equation}
    \hfill
    \hat{y}_{ij} = O_i \cdot p_{ij} \equiv O_i \frac{e^{f(x_{ij})}}{\sum^{N}_{k=1} e^{f(x_{ik})}}
    \hfill
    \label{eq:y_pblack}
\end{equation}

\noindent
where $O_i$ represents the total number of ships departing from source port $i$.
$p_{ij}$ is the probability of ships traveling from source port $i$ to the destination region $j$ among all the possible destinations $1 \sim N$, and $f(x_{ij})$ is the model output of feature vector $x_{ij}$.

\noindent\textit{\textbf{Loss function:}}
We used the cross-entropy loss function for the model optimization process, defined as:
\begin{equation}
    \hfill
    L\left(\hat{y}_{ij}, y_{ij}\right)
    = -\sum^{M}_{i=1}\sum^{N}_{j=1} y_{ij} \cdot \ln \left(\frac{e^{f(x_{ij})}}{\sum^{N}_{k=1}e^{f(x_{ik})}}\right)
    = -\sum^{M}_{i=1}\sum^{N}_{j=1} y_{ij} \cdot \ln \left(\frac{\hat{y}_{ij}}{O_i}\right)
    \hfill
    \label{eq:loss_func}
\end{equation}

\noindent
The function presents the total loss between the predicted flows $\hat{y}_{ij}$, and the real flows $y_{ij}$ for all the $N$ destination regions from $M$ source ports.
The {\it log-softmax} function is applied to the model output $f(x_{ij})$, and the loss function in terms of $p_{ij}$ is obtained by replacing the log-term by Equation~\ref{eq:y_pblack} divided by $O_i$.

\noindent\textit{\textbf{Evaluation metric:}}
The \textit{Common Part of Commuters} (CPC)~\cite{Luca2021SurveyDeep, LENORMAND2016158, Barbosa2018HumanMobility} is designed to measure the similarity between two sets of data, which could represent various aspects such as the volume of commuters, traffic, or trade between different locations.
The metric calculates how much overlap there is between the predicted values $\hat{y}_{ij}$ and the actual values $y_{ij}$.
In Equation~\ref{eq:cpc}, $M$ represents the number of source ports and $N$ the number of destination regions.
The values $\hat{y}_{ij}$ and $y_{ij}$ correspond to the flow of vessels from source port $i$ to destination region $j$, based on a model's prediction and the actual observed values, respectively.
In this case, a high value of the CPC means a significant overlap between the predicted and actual datasets.
Specifically, it would indicate that the predictions accurately capture the true data distribution patterns, with most predictive quantities closely matching the actual quantities.
Contrarily, a low value of the CPC would suggest that there is little overlap between the predictions and the actual data, indicating that the model's predictions diverge significantly from the observed data, which could be due to underprediction or overprediction in various parts or a general misalignment of the model with the reality.
Accordingly, CPC considers the minimum common value between the predicted and actual data for each pair of source and destination ports, measuring the intersection over the values union:
\begin{equation}
    \hfill
    CPC(\hat{y}_{ij}, y_{ij}) = \sum^{M}_{i=1}\frac{2\sum^{N}_{j=1} \min(\hat{y}_{ij}, y_{ij})}{\sum^{N}_{j=1} \hat{y}_{ij} + \sum^{N}_{j=1} y_{ij} }
    \hfill
    \label{eq:cpc}
\end{equation}

To evaluate the model's performance comprehensively, besides the CPC, we included the Normalized Root Mean Square Error ($NRMSE$) --- lower is better --- and Pearson Correlation Coefficients ($Corr.$) --- higher is better --- to measure the normalized errors and the correlation between the predictions and observations, and to provide a multi-faceted assessment of the model's accuracy and reliability, defined as:
\begin{equation}
    \hfill
    NRMSE(\hat{y}_{ij}, y_{ij}) = \sum^{M}_{i=1} \frac{\sqrt{\frac{1}{N}\sum_{j=1}^{N}(y_{ij} - \hat{y}_{ij})^2}}{\max(y_{ij}) - \min(y_{ij})}
    \hfill
    \label{eq:nrmse}
\end{equation}
\begin{equation}
    \hfill
    Corr.(\hat{y}_{ij}, y_{ij}) = \sum^{M}_{i=1} \frac{\sum_{j=1}^{N} (y_{ij} - \overline{y}_{ij})(\hat{y}_{ij} - \overline{\hat{y}}_{ij})}{\sqrt{\sum_{j=1}^{N} (y_{ij} - \overline{y}_{ij})^2 \sum_{j=1}^{N} (\hat{y}_{ij} - \overline{\hat{y}}_{ij})^2}}
    \hfill
    \label{eq:pearson_corr}
\end{equation}

\subsubsection*{Model Framework}
Our proposed Transformer Gravity model is composed of three main components: 
(1) the input embedding layer, which maps the input feature vectors to a higher-dimensional space that is compatible with the Transformer architecture;
(2) the multilayer Transformer encoder, which involves the self-attention and feed-forward blocks that process the embeddings to capture complex relationships between input features; and,
(3) the output linear layer, which maps the processed embeddings to the target flow predictions, computes loss and CPC and performs backpropagation based on the loss values.
The input embedding layer (1) helps us organize and make sense of that data by putting it into a format that is easier for a machine to understand. 
This format is represented as vectors of numerical values, where each dimension captures specific input data characteristics and is fed into subsequent layers of the model for further processing until achieving the output layer. The output represents the predicted value, which, in our case, is the inflow or outflow of vessels between a pair of ports.
The multilayer Transformer encoder (2) looks at all the different parts of information organized in the first component and finds out how they relate. 
Transformers are known for spotting patterns and connections between different parts of the training data. This helps our model understand the complex relationships between various aspects of the shipping data, like which routes are busiest, or ports are most important.
The output linear layer (3) helps us make predictions based on the patterns and connections found by the previous layer. This layer will predict how much shipping traffic will be in a particular area. If the predictions are not entirely correct, this layer also helps us learn from our mistakes and improve our future guesses due to the backpropagation mechanism. Backpropagation is a process where the model adjusts its internal parameters based on the difference between its predictions and the actual outcomes, gradually refining its understanding of the data and improving its predictive power over time.
Figure~\ref{fig:model_pipeline} presents the model pipeline using two stacked Transformers modules and provides a glance at the layer's relationships.

\noindent\textit{\textbf{Linear Embedding.}}
The embedding layer takes the input sample, which is a sequence of feature vectors represented as $\{x_{i1}, x_{i2}, \ldots, x_{iN}\}$.
It then maps each vector into a higher-dimensional space using a linear transformation that involves a weights matrix and a bias vector.
The result of this transformation is the feature embedding $\textbf{z}_0$ for each vector $x_{ij}$, which can be obtained following the subsequent calculation:
\begin{equation}
    \hfill
        \textbf{z}_0 = x_{ij} \cdot {W_0}^{\top} + b_0, 
        ~ x_{ij} \in \mathbb{R}^{1 \times n}, 
        ~ W_0 \in \mathbb{R}^{d \times n}, 
        ~ b_0 \in \mathbb{R}^{1 \times d}
    \hfill
    \label{eq:feature_embedding}
\end{equation}

\noindent
The input vector $x_{ij}$ with 10 features is represented by $W_0$ (the weight matrix) and $b_0$ (the bias vector).
This input vector is then embedded into a 64-dimensional space, resulting in an embedded output $\textbf{z}_0 \in \mathbb{R}^{1 \times d}$.
Subsequently, the embedded output is passed to the multi-head attention encoder layers as the input.

\noindent\textit{\textbf{Multi-Head Attention.}}
A multi-head attention encoder comprises a multi-head self-attention mechanism and a feed-forward network, followed by layer normalization (as illustrated in Figure~\ref{fig:model_pipeline}).
Within each self-attention head, the input $\textbf{z}_0$ is transformed into queries $Q_h$, keys $K_h$ and values $V_h$ using the weight matrices $W_Q$, $W_K$ and $W_V$, respectively.
Self-attention then calculates $Head_h = softmax\left(\frac{Q_h \cdot  {K_h}^{\top}}{\sqrt{d_k}}\right) \cdot V_h$, where $d_k = \frac{d}{h}$ is the dimension of the queries $Q_h$ and keys $K_h$ and is used to scale the product $Q_h \cdot {K_h}^{\top}$.
Multi-head attention combines all heads and linearly transforms the concatenation to produce $\textbf{z}_1 \in \mathbb{R}^{1 \times d}$:
\begin{equation}
    \hfill
    \textbf{z}_1 = Concat(Head_1, \ldots, Head_h) \cdot W_C,
    ~~ W_C \in \mathbb{R}^{h d_v \times d}
    \hfill
    \label{eq:multi-head}
\end{equation}

\noindent
Our experiment defines the number of heads as $h=2$, and the heads run operations in parallel.

\noindent\textit{\textbf{Layer Normalization.}}
After the multi-head attention layer, a dropout layer randomly sets a certain percentage of elements to 0.
The dropout ratio $p$ is set to $0.1$ as per Equation~\ref{eq:layer_norm}.
Next, a skip connection is applied to add the input features $\textbf{z}_0$ to the output of the dropout layer $\textbf{z}_{dropout}$ before the self-attention block.
This helps retain the input features' information and prevents vanishing gradients during backpropagation.
The output of this connection, $\textbf{z}_{skip}$, is then normalized using Equation~\ref{eq:layer_norm}, where $\mu$ is the mean and $\sigma$ is the standard deviation of $\textbf{z}_{skip}$ with a small bias.
The affine parameters $\alpha$ and $\beta$ are initialized as 1 and 0, respectively, and can be optimized during training.
\begin{equation}
    \hfill
    \begin{aligned}
        \textbf{z}_{skip} &= \textbf{z}_0 + \textbf{z}_{dropout} \equiv \textbf{z}_0 + Dropout(\textbf{z}_1, p) \\
        \textbf{z}_2 &= LayerNorm(\textbf{z}_{skip}) \equiv \frac{\textbf{z}_{skip}-\mu}{\sigma} \times \alpha + \beta
    \end{aligned}
    \hfill
    \label{eq:layer_norm}
\end{equation}

\noindent\textit{\textbf{Feed-Forward Network.}}
After the multi-head attention block processes the input, the resulting output is fed into a feed-forward neural network composed of an MLP structure.
The connectivity of each layer in the feed-forward block's structure is illustrated in Figure~\ref{fig:model_pipeline}.
We formulate the output vectors from the layers using corresponding weight updates in Equation~\ref{eq:feed-forward}.
Similar to $\textbf{z}_{skip}$, a skip connection adds the vector $\textbf{z}_2$ to the output $\textbf{z}_4$ to preserve information from the self-attention block.
\begin{equation}
    \hfill
    \begin{aligned}
        \textbf{z}_3 &= Dropout \left( ReLU( \textbf{z}_2 \cdot W^{\top}_1 + b_1), p\right) \\
        \textbf{z}_4 &= Dropout \left( (\textbf{z}_3 \cdot W^{\top}_2 + b_2), p \right) \\
        \textbf{z}_5 &= LayerNorm(\textbf{z}_4 + \textbf{z}_2)
        \label{eq:feed-forward}
    \end{aligned}
    \hfill
\end{equation}

\noindent\textit{\textbf{Training and Optimization.}}
Our Transformer Gravity model has three transformer encoder layers stacked together to capture complex input embedding dependencies. Still, the number of stacked layers can be changed to match different requirements and needs.
The output value $\textbf{z}_5$ is obtained by passing the output of Equation~\ref{eq:feed-forward} through a second and later third multi-head attention and feed-forward network block.
The output value of the model is denoted as $f(x_{ij})$, and it produces a sequence of output values for a single data sample with a length of $N$.
This sequence is then applied to a \textit{softmax} function to produce probabilities $\{p_{i1}, p_{i2}, \ldots, p_{ij},\ldots, p_{iN}\}$ for $N$ classes.
The predicted flow sizes $\{ \hat{y}_{i1}, \hat{y}_{i2}, \ldots, \hat{y}_{ij},\ldots, \hat{y}_{iN} \}$ for each destination are obtained by multiplying these probabilities with the total outflows $O_i$ from the source port, as given in Equation~\ref{eq:y_pblack}.
The loss for every sample is computed using Equation~\ref{eq:loss_func}, and these losses are collected to derive the total loss.
The model's parameters are updated with each loss by processing a single sample.
The summed CPC across all samples is calculated using Equation~\ref{eq:cpc}.
After each training epoch, the summed CPC is divided by the number of samples $M$ ({\it i.e.}, the number of source ports) to obtain the average CPC of that epoch.
We used the Adam optimizer with $L_2$ regularization on the weights during training. We reduced the learning rate by a factor of 0.1 when there was no improvement in the validation CPC after 10 epochs.
We used early stopping when there was no improvement with the validation CPC after 20 epochs to prevent overfitting and improve training efficiency.

\bibliography{reference}

\section*{Acknowledgments}
This research was partially funded by the Natural Sciences and Engineering Research Council of Canada (NSERC), Grant number RGPIN-2022-03909; Sao Paulo State Research Foundation (FAPESP), Grant number 2016/17078-0 and 2022/12374-1; the Institute for Big Data Analytics (IBDA) and the Ocean Frontier Institute (OFI) at Dalhousie University, Halifax - NS, Canada; and further funded by the Canada First Research Excellence Fund (CFREF), the Canadian Foundation for Innovation MERIDIAN cyberinfrastructure.
We thank all funding sources and {\it Spire} for providing the vessel trajectory dataset.
We also thank Sarah Bailey, from the Great Lakes Laboratory for Fisheries and Aquatic Sciences, Fisheries and Oceans, Dartmouth – NS, Canada, for reviewing and sharing her thoughts about this paper.

\section*{Author contributions statement}
Conceptualization, R.S. and G.S.;
methodology, R.S., G.S. and A.S.;
validation, R.S., G.S., and A.S.;
formal analysis, R.S., and G.S.;
data modeling, R.S. and G.S.;
data curation, R.S., G.S., and A.S.;
writing --- original draft preparation, R.S. and G.S.;
writing --- review and editing, all authors;
visualization, R.S.;
supervision, G.S., A.S, R.P., and S.M.;
project administration, G.S. and A.S.;
funding acquisition, A.S., S.M., R.P., and G.S.;
All authors have read and agreed to the published version of the manuscript.

\section*{Data Availability}
The research we describe in this paper has used data acquired from Spire under MERIDIAN's fair-use and non-disclose data license, which is research-intended high-resolution data that is hosted at the Institute for Big Data Analytics (IBDA) from Dalhousie University. Due to licensing agreements, we cannot share the raw data, but we can provide a trained model instead; contact the corresponding author for assistance.

\section*{Competing interests}
The authors declare no competing interests.
\end{document}